\documentclass[twoside,11pt]{article}

\usepackage{jmlr2e}
\usepackage{algorithm}
\usepackage{xcolor}
\usepackage{amsmath}
\usepackage[utf8]{inputenc}
\usepackage{algorithmic}
\usepackage{amsfonts}
\usepackage[dont-mess-around]{fnpct}
\newcommand{\x}{\mathbf{x}}
\usepackage{multirow}
\newcommand{\subf}[2]{%
  {\small\begin{tabular}[t]{@{}c@{}}
  #1\\#2
  \end{tabular}}%
}

\jmlrheading{1}{2000}{1-48}{4/00}{10/00}{Marina Meil\u{a} and Michael I. Jordan}

\firstpageno{1}

\begin{document}

\title{Mitigating shortage of labeled data using clustering-based active learning with diversity exploration }

\author{\name Xuyang Yan \email xyan@@aggies.ncat.edu \\
       \addr Dept. of ECE, North Carolina A \& T State University, NC, USA.
       \AND
       \name Shabnam Nazmi \email snazmi@aggies.ncat.edu \\
       \addr Dept. of ECE, North Carolina A \& T State University, NC, USA.
       \AND
       \name Biniam Gebru \email btgebru@aggies.ncat.edu \\
       \addr Dept. of ECE, North Carolina A \& T State University, NC, USA.
       \AND 
       \name Mohd Anwar \email manwar@ncat.edu  \\
       \addr Dept. of Computer Science, North Carolina A \& T State University, NC, USA.
      \AND
      \name Abdollah Homaifar \email homaifar@@ncat.edu \\
      \addr Dept. of ECE, North Carolina A \& T State University, NC, USA.
      \AND 
      \name Mrinmoy Sarkar \email msarkar@aggies.ncat.edu   \\
      \addr Dept. of ECE, North Carolina A \& T State University, NC, USA.
      \AND 
      \name Kishor Datta Gupta \email kgupta@cau.edu \\
      \addr Dept. of Computer Science, Clark Altanta University, GA, USA\\}


\maketitle

\begin{abstract}
In this paper, we proposed a new clustering-based active learning framework, namely Active Learning using a Clustering-based Sampling (ALCS), to address the shortage of labeled data. ALCS employs a density-based clustering approach to explore the cluster structure from the data without requiring exhaustive parameter tuning. 
A bi-cluster boundary-based sample query procedure is introduced to improve the learning performance for classifying highly overlapped classes. Additionally, we developed an effective diversity exploration strategy to address the redundancy among queried samples. Our experimental results justified the efficacy of the ALCS approach.
\end{abstract}


\section{Introduction}
Active learning (AL) \citep{lewis1994sequential} approaches have been proposed to address the scarcity of associated labels and reduce the annotation costs for predictive modeling. As an important branch of AL, clustering-based AL methods are proposed to explore the \textit{representativeness} of samples and they have shown reasonable success \citep{huang2010active,wang2017active,wang2018active,wang2020active}. 
In clustering-based AL approaches, samples are assumed to share the same class label within the same cluster so that AL is conducted by querying the representative samples from those clusters \citep{dasgupta2008hierarchical}.
\paragraph{Challenges.} Despite the success of the existing clustering-based AL methods, three major challenges are identified as follows: the performance of existing clustering-based AL methods strongly depends on the selection of clustering parameters; the existing boundary-based selection strategy primarily queries labels for the farthest samples in each cluster without considering neighboring clusters \citep{wang2018active,wang2020active}; limited efforts are made on clustering-based AL methods to consider the diversity among queried samples \citep{wang2017incorporating,kee2018query,wang2020active,xiao2020efficient}.   \\
\indent In light of these challenges, a clustering-based AL method, namely \textbf{AL} utilizing a \textbf{C}lustering-based \textbf{S}ampling (ALCS)\footnote{This work has been published in the Springer Applied Intelligence: https://doi.org/10.1007/s10489-021-03139-y.}, is proposed. ALCS adopts a density-based clustering technique, namely fitness proportionate sharing clustering (FPS-clustering) \citep{yan2017novel}, to relax the dependency on clustering parameter optimization. We develop a new bi-cluster boundary-based selection procedure to improve the learning performance of ALCS in datasets with highly overlapped classes. Furthermore, an effective diversity exploration strategy is introduced to reduce the redundancy among active queried samples.

\section{Proposed methodology} \label{section:4}
\begin{figure}[h]
    \centering
    \includegraphics[width = 2.6in,height=2.7in]{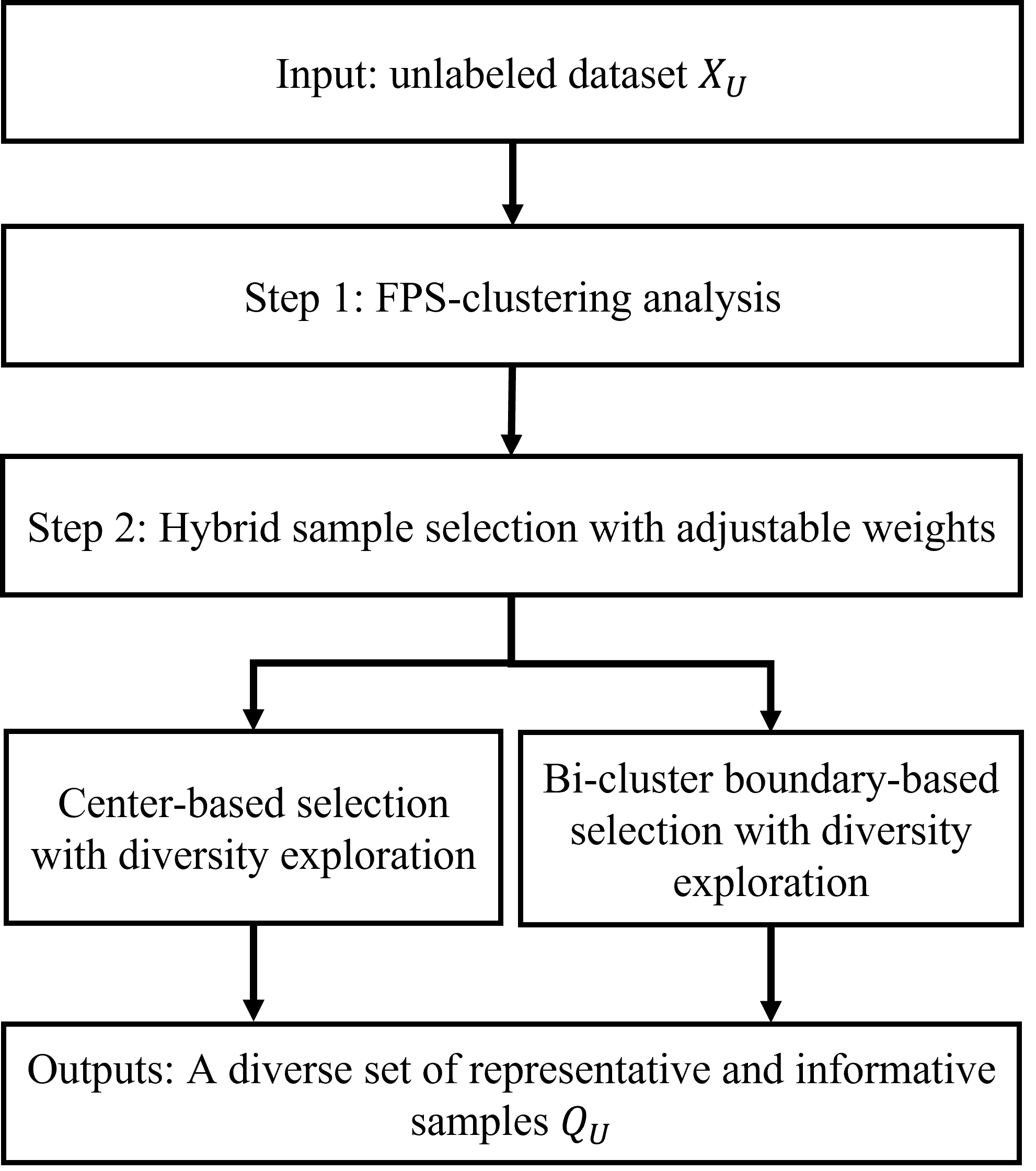}
    \centering
    \caption{A workflow of the proposed clustering-based AL framework.}
    \label{fig:complete procedure}
\end{figure}

In this section, the ALCS technique is discussed in terms of its two main components: (i) clustering; and  (ii) distance-based instance selection with diversity exploration. Figure \ref{fig:complete procedure} provides an overview of the ALCS technique and details are discussed below. 
\subsection{Clustering}
To effectively alleviate the exhaustive parameter tuning issue, the FPS-clustering algorithm is employed to discover the cluster information as the first step of the ALCS technique. The FPS-clustering algorithm takes the unlabeled dataset $X_{U}$ as the input and then outputs a set of clusters and the corresponding cluster information $\Omega$, which is expressed as follows:
\begin{equation}
     \Omega  = \{(C_{i},\mathbf{d}(C_{i}))| i=1,...,c\}, \label{eq:4}
\end{equation}
\begin{equation}
    \mathbf{d}(C_{i}) = \{d(\x_{i}^{j},C_{i})|j=1,...,|C_{i}|\}. \label{eq:5}
\end{equation}
Here, $C_{i}$ refers to the center of the $i^{th}$ cluster, $c$ denotes the total number of discovered clusters, and $d(\x_{i}^{j},C_{i})$ is the distance from sample $\x_{i}^{j}$ to its respective cluster center $C_i$. 
The cardinality $|C_{i}|$ denotes the number of samples that belong to $C_{i}$. 
\subsection{Distance-based sample selection}
\paragraph{Hybrid sample selection strategy.}
After the clustering procedure, ALCS employs a novel hybrid sample selection strategy for active label query. Let the number of queried samples from the $i_{th}$ cluster be $n_{q_{i}}$. The bi-cluster boundary-based selection step takes $\lfloor n_{q_{i}}\times \rho_{i}\rceil$ samples from cluster $i$ as $Q_{boundary}$ where $\rho_{i}$ denotes the sampling weight from the boundary of two adjacent clusters. Accordingly, the center-based selection policy chooses the remaining $\lfloor n_{q_{i}}\times (1 - \rho_{i})\rceil$ samples from the center region as $Q_{center}$. The value of $\rho_{i}$ ranges from zero to one. Algorithm \ref{algorithm:2} summarizes the hybrid sample selection procedure.
\begin{algorithm}[tb]
\scriptsize
\caption{The hybrid sample selection strategy}
\label{algorithm:2}
\textbf{Input}: $\Omega$, $n_{q}$, $\rho_{i}$\\
\textbf{Parameter}: $Q_{center}$, $Q_{boundary}$, and $n_{q_{i}}$\\
\textbf{Output}: The set of queried samples $Q_{U}$
\begin{algorithmic}[1] 
\STATE $Q_{U} = \emptyset$
\FOR{i = 1 to $n_{C}$}
\STATE Calculate the number of queries for cluster $i$: $n_{q_{i}} = \lfloor \frac{|C_{i}|}{n_{U}}\times n_{q} \rceil$
\STATE Perform the center-based query with diversity exploration to obtain $Q_{center}$ using equation \ref{eq:6}.
\STATE Perform the bi-cluster boundary-based query with diversity exploration to obtain $Q_{boundary}$ using equation \ref{eq:9}
\STATE $Q_{U} = Q_{U} \cup \{Q_{center} \cup Q_{boundary}\}$
\ENDFOR
\STATE \textbf{return} $Q_{U}$
\end{algorithmic}
\end{algorithm}
\paragraph{Center-based sample selection.}
For center-based selection, the query priority of each sample is computed in terms of \textbf{C}luster \textbf{R}epresentativeness (CR). Let $CR(*)$ and $P(*)$ be the cluster representativeness and query priority functions for clustered samples, respectively. For center-based selection, the query priority of $x^{j}_{i}$ is calculated below.
\begin{equation} \label{eq:6}
P(x^{j}_{i}) = CR(x^{j}_{i}),
\end{equation}
and 
\begin{equation} \label{eq:7}
    CR(x^{j}_{i}) = \frac{1}{1+e^{d(x^{j}_{i}, C_{i})}}.
\end{equation}
Where $d(x^{j}_{i}, C_{i})$ refers to the distance from $x^{j}_{i}$ to $C_{i}$. From equation \ref{eq:7}, the representativeness of each sample is inversely proportional to its distance to $C_{i}$ and samples that are close to the cluster center have higher representativeness. 
\paragraph{Bi-cluster boundary-based sample selection.}
We propose an effective bi-clusters boundary-based selection strategy to identify the most uncertain samples using the distance to their assigned cluster center and neighboring cluster center. This strategy utilizes the law of cosines to query the most informative samples from the cross-boundary region with two adjacent cluster centers. Assume the candidate bi-boundary sample in the $i^{th}$ cluster is $x^{j}_{CB_{i}}$ and the candidate set is $CB = \{x^{j}_{CB_{i}}|j=1,...,\frac{|C_{i}|}{2}\}$. The two adjacent cluster centers are denoted as $NC_{1}$ and $NC_{2}$. The query priority of $x^{j}_{CB_{i}}$ is calculated using the \textbf{C}luster \textbf{U}ncertainty (\textit{CU}), which is expressed as follows:
\begin{equation}
    P(x^{j}_{CB_{i}}) = CU(x^{j}_{CB_{i}}), \label{eq:8} 
\end{equation}
and
\begin{equation}
    CU(x^{j}_{CB_{i}}) = \frac{1}{1+e^{\frac{d_1+d_2}{d_{ref_{1}}+d_{ref_{2}}}}}. \label{eq:9} 
\end{equation}
Where $d_{ref_{1}}$ and $d_{ref_{2}}$ denote the distance from $C_{i}$ to its two neighboring cluster centers. Here, $d_1$ refers to the distance from $x^{j}_{CB_{i}}$ to $NC_{1}$ and $d_2$ refers to the distance from $x^{j}_{CB_{i}}$ to $NC_{2}$, respectively. Detailed derivations can be found in \citep{yan2022clustering}. According to equation \ref{eq:9}, a sample is considered to have higher uncertainty when it has a larger $CU$.
\paragraph{Diversity exploration.} \label{diversity}
We developed a diversity exploration strategy based on fitness proportionate niching (FPN) \citep{workineh2012fitness} to guide the search for informative and representative samples. 
Let $X_{C_{i}}$ be a set of samples that belongs to $C_{i}$, the query priority function is expressed as follows.
\begin{equation} \label{eq:15}
    P(X_{C_{i}}) = 
    \begin{cases}
    CR(X_{C_{i}}), & \text{ query from centers;}\\
    CU(X_{C_{i}}), & \text{ query from boundaries.}
    \end{cases}
\end{equation}
Based on the priority function, the proposed diversity exploration procedure aims to decompose $X_{C_{i}}$ into a number of small niches and query a set of diverse samples from different niches. During the sample selection procedure, the sample with the highest query priority is inserted into the queried sample set initially. Then, a niche can be formed by a set of samples in the neighborhood of this sample and a priority sharing strategy is employed to decrease the query priorities of other samples in the niche. The average distance for all $k$-nearest-neighbor graphs within a cluster is used as the neighborhood radius. As a rule of thumb, we set the value of $k$ to be the square root of cluster size. Assume $n^{j}_{i}$ and $X_{n^{j}_{i}}$ denote the $j^{th}$ niche in $C_{i}$ and a set of samples belong to $n^{j}_{i}$, respectively. Equation \ref{eq:16} describes the priority sharing function.
\begin{equation}\label{eq:16}
    P(X_{n^{j}_{i}}) = \frac{P(X_{n^{j}_{i}})}{\sum{P(X_{n^{j}_{i}})}}, x_{i} \in n^{j}_{i}.
\end{equation} 
From equation \ref{eq:16}, samples from the same niche will have relatively low priorities during the next sample query stage. Consequently, it guarantees to query more diverse samples from each cluster. 
\section{Experiments and results} \label{section:5}
\begin{figure*}
\centering
\begin{tabular}{cc}
\subf{\includegraphics[width=55mm]{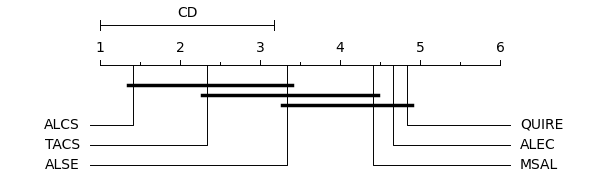}}
     {(a) $Acc$}
&
\subf{\includegraphics[width=55mm]{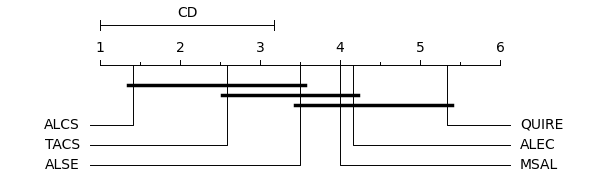}}
     {(b) $F_{mac}$}
\end{tabular}
\caption{Comparison of ALCS against other clustering-based AL methods with the Nemenyi test with $\alpha=0.05$.}
\label{fig:cd clustering based}
\end{figure*}

\paragraph{Datasets and compared methods.}
Twelve benchmark datasets from \citep{Dua:2017} are used in the experiments. 
We compared ALCS with five state-of-the-art clustering-based AL methods, including QUIRE \citep{huang2010active}, ALEC \citep{wang2017active}, active learning through multi-standard optimization (MSAL) \citep{wang2019active}, active learning through label error statistical (ALSE) \citep{wang2020active}, and three-way active learning through clustering selection (TACS) \citep{min2020three}. The implementation of the ALCS method, using python, is available at a Github repository \footnote{https://github.com/XuyangAbert/ALCS}.
\paragraph{Result discussions.}
\indent Table \ref{table_res2} compares the performance of ALCS with five clustering-based AL approaches using $10\%$ labeled data. It is observed that ALCS provides better performance in most datasets, and it has the highest average ranks for both \textit{Acc} and $F_{mac}$. In Australian, Aggregation, Spambase, Waveforms, Electricity, Penbased, GasSensor, and MNIST datasets, ALCS outperforms the other five clustering-based AL methods on both Acc and Fmac metrics. These results imply the efficacy of ALCS in handling datasets with highly overlapped classes.
The Nemenyi post-hoc test is performed with a significance level of $0.05$ and results are shown in Figure \ref{fig:cd clustering based}. Figure \ref{fig:cd clustering based} displays that ALCS is statistically better than ALEC, QUIRE, and MSAL methods in terms of $Acc$ and $F_{mac}$. On the other hand, ALCS presented statistically comparable performance with TACS and ALSE. 
\begin{table}[h]
\caption{Performance comparison of ALCS and five clustering-based AL methods.}\label{table_res2}
\centering
\resizebox{0.65\columnwidth}{!}{
\begin{tabular}{l|ccccccc}
\hline
Dataset        &  Metrics         & ALEC       & QUIRE      & MSAL       & ALSE       & TACS      & ALCS      \\ \hline
R15            & \textit{Acc}         & 84.58(6) & \textbf{99.26(1)} & 99.14(2) & 86.27(5) & 98.45(4)  & 99.07(3) \\
                & $F_{mac}$ & 84.09(6) & \textbf{99.21(1)} & 98.27(3) & 83.94(5) & 97.66(4) & 99.06(2) \\
Australian     & \textit{Acc}         & 80.80(5) & 81.29(4) & 68.78(6) & 81.38(3) & 82.08(2) & \textbf{83.31(1)} \\
                & $F_{mac}$ & 79.71(6)  & 80.87(3) & 68.69(6) & 80.82(4) & 80.92(2) & \textbf{83.06(1)} \\
Aggregation     & \textit{Acc}         & 91.06(5) & 71.01(6) & 91.25(4) & 91.91(3) & 92.74(2) & \textbf{99.43(1)} \\
                & $F_{mac}$ & 76.86(5) & 44.21(6) & 76.92(4) & 77.63(3) & 78.15(2) & \textbf{99.09(1)} \\
Vehicle        & \textit{Acc}         & 46.11(6) & 53.23(3) & 48.92(4) & 46.39(5) & 53.45(2) & \textbf{54.74(1)} \\
                & $F_{mac}$ & 54.66(3) & 49.52(6) & \textbf{55.12(1)} & 52.37(5) & 54.83(2) & 54.30(4)  \\
Spambase        & \textit{Acc}         & 76.48(4) & 75.73(5) & 75.32(6) & 76.57(3) & 79.58(2) & \textbf{81.54(1)} \\
                & $F_{mac}$ & 75.85(3) & 74.79(6) & 75.87(5) & 75.46(4) & 80.28(2) & \textbf{80.92(1)} \\
Waveforms      & \textit{Acc}         & 75.42(5) & 75.87(4) & 75.32(6) & 76.89(3) & 78.17(1) & 76.66(2) \\
                & $F_{mac}$ & 75.84(3) & 74.91(6) & 75.47(4) & 75.12(5) & 76.52(2) & \textbf{76.62(1)} \\
Electricity    & \textit{Acc}        & 82.81(5) & 82.48(6) & 83.01(3) & 83.22(2) & 82.88(4) & \textbf{85.34(1)} \\
                & $F_{mac}$ & 80.47(3) & 79.89(6) & 80.44(5) & 80.83(2) & 80.56(4) & \textbf{83.76(1)} \\
DLA0.01         & \textit{Acc}         & 86.27(5) & 72.14(6) & 92.48(4) & 93.18(3) & \textbf{99.22(1)} & 93.61(2) \\
                & $F_{mac}$ & 86.28(5) & 72.51(6) & 86.98(4) & 87.15(3) & \textbf{97.98(1)} & 88.26(2) \\
Penbased       & \textit{Acc}         & 87.94(5) & 82.74(6) & 89.48(3) & 88.13(4) & 91.24(2) & \textbf{94.80(1)}  \\
                & $F_{mac}$ & 86.98(5) & 72.68(6) & 88.04(4) & 89.01(3) & 91.03(2) & \textbf{94.76(1)} \\
GasSensor & \textit{Acc}         & 64.94(5) & 64.40(6) & 65.79(4) & 66.44(3) & 66.88(2) & \textbf{72.81(1)} \\
                & $F_{mac}$ & 61.95(5) & 60.60(6) & 62.84(4) & 63.74(3) & 64.25(2) & \textbf{71.55(1)} \\
DCCC           & \textit{Acc}         & \textbf{76.88(1)} & 75.15(5) & 74.85(6) & 75.26(4) & 75.45(3) & 76.43(2) \\
                & $F_{mac}$ & 54.16(4) & 44.21(6) & 49.56(5) & 57.35(2) & 54.95(3) & \textbf{60.84(1)} \\
MNIST           & \textit{Acc}         & 87.58(4) & 84.52(6) & 87.12(5) & 88.45(2) & 87.75(3) & \textbf{91.83(1)} \\
                & $F_{mac}$ & 87.15(2) & 83.48(6) & 86.54(4) & 86.81(3) & 84.07(5) & \textbf{91.79(1)}\\
\hline
    \multirow{2}{*}{Avg. ranks} & \textit{Acc} & 4.67 & 4.83 & 4.42 & 3.33 & 2.33 & \textbf{1.42}\\
    & $F_{mac}$ & 4.17 & 5.33 & 4.08 & 3.50 & 2.58 & \textbf{1.41}\\
    \hline
\end{tabular}}
\end{table}

\section{Concluding remarks} \label{section:6}
In this paper, we presented a novel active learning framework using clustering-based sampling to handle the shortage of prior label information. It utilizes the FPS-clustering procedure to explore the structure of unlabeled data without exhaustive parameter tuning. A new distance-based sample selection procedure with an effective diversity exploration strategy was developed to enhance the quality of queried labels. Experimental results established that ALCS provided better or comparable performance than the five clustering-based AL approaches without tuning the clustering parameters. 

\paragraph{Merits, Limitations \& Future work.} As a new clustering-based AL framework, ALCS effectively handles the dependency on clustering parameters and offers a promising solution to improve the diversity among queried labels. Moreover, the bi-cluster boundary selection strategy is designed to enhance the learning performance in datasets with highly overlapped classes. Limitations of the ALCS can be summarized from two aspects: (i) the imbalance among different class distributions is not considered in ALCS; and (ii) ALCS is currently limited to offline AL problems;  Therefore, our future work will focus on addressing these two limitations of the ALCS framework. 



\acks{We would like to acknowledge support for this project
from the Air Force Research Laboratory and the Office of the Secretary of Defense under agreement number (FA8750-15-2-0116). This work
is also partially supported by National Science Foundation under grant number
(2000320).}










\vskip 0.2in
\bibliography{sample}

\end{document}